\documentclass{article}
\usepackage{spconf,amsmath,amssymb,graphicx,algorithm,algorithmic,cite,bm,stfloats}
\usepackage[resetlabels]{multibib}

\usepackage{xcolor}
\usepackage[normalem]{ulem}
\useunder{\uline}{\ul}{}
\newcites{sec}{Reference}

\def\LL{{\mathcal L}}
\def\thet{{\bm \theta}}

\title{Unrolled Graph Learning for Multi-Agent Collaboration}

\name{Enpei Zhang\textsuperscript{\footnotesize{\rm 1}}, Shuo Tang\textsuperscript{\footnotesize{\rm 1,2}}, Xiaowen Dong\textsuperscript{\footnotesize{\rm 3}}, Siheng Chen\textsuperscript{\footnotesize{\rm 1,2}}, Yanfeng Wang\textsuperscript{\footnotesize{\rm 2,1}}}
\address{\textsuperscript{\footnotesize{\rm 1}}Shanghai Jiao Tong University,
\textsuperscript{\footnotesize{\rm 2}}Shanghai AI Laboratory,
\textsuperscript{\footnotesize{\rm 3}}University of Oxford}

\begin{document}
\maketitle

\begin{abstract}
Multi-agent learning has gained increasing attention to tackle distributed machine learning scenarios under constrictions of data exchanging. However, existing multi-agent learning models usually consider data fusion under fixed and compulsory collaborative relations among agents, which is not as flexible and autonomous as human collaboration. To fill this gap, we propose a distributed multi-agent learning model inspired by human collaboration, in which the agents can autonomously detect suitable collaborators and refer to collaborators’ model for better performance.
To implement such adaptive collaboration, we use a collaboration graph to indicate the pairwise collaborative relation. 
The collaboration graph can be obtained by graph learning techniques based on model similarity between different agents.
Since model similarity can not be formulated by a fixed graphical optimization, we design a graph learning network by unrolling, which can learn underlying similar features among potential collaborators.
By testing on both regression and classification tasks, we validate that our proposed collaboration model can figure out accurate collaborative relationship and greatly improve agents' learning performance.
\end{abstract}


\section{Introduction}
\label{sec:intro}
Collaboration is an ancient and stealthy wisdom in nature. When observing and understanding the world, each individual has a certain bias due to the limited field of view. The observation and cognition would be more holistic and robust when a group of individuals could collaborate and share information~\cite{almaatouq2020adaptive}. Motivated by this, multi-agent collaborative learning is emerging~\cite{panait2005cooperative,2103.00710,WarnatHerresthal2021SwarmLF}. Currently, most collaborative models are implemented by either a centralized setting or distributed settings with predefined, fixed data-sharing topology~\cite{9084352,2103.00710,WarnatHerresthal2021SwarmLF,rieke2020future,lalitha2018fully,minami2020knowledge,jiang2017collaborative,1709.05412}. However, as the role model of collaborative system, human collaboration is fully distributed and autonomous, where people can adaptively choose proper collaborators and refer to others' information to achieve better local task-solving ability~\cite{almaatouq2020adaptive}. This is much more robust, flexible and effective than centralized setting or predefined collaboration.

\begin{figure}[t]
    \centering
    \includegraphics[width=0.98\linewidth]{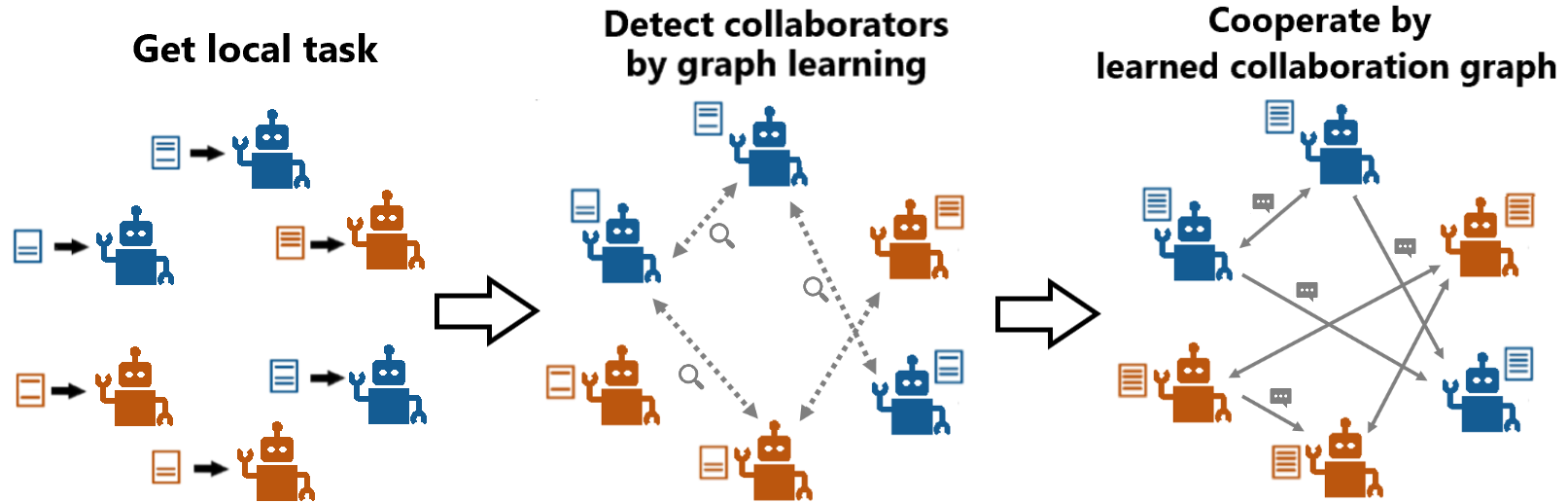}
    \vspace{-0.1cm}
    \caption{Human-like distributed collaborative learning.}
    \label{fig:collab}
    \vspace{-0.2cm}
\end{figure}

To fill this gap, we consider a human-like collaborative learning mechanism, see \textbf{Fig.~\ref{fig:collab}}. In this setting, the agents are expected to autonomously find proper collaborators and refer to others' models, which grants adaptive knowledge transfer pattern and preserves a personalized learning scheme. Following this spirit, we formulate a novel mathematical framework of a distributed and autonomous multi-agent learning. In our framework, each agent optimizes its local task by alternatively updating its local model and collaboration relationships with other agents who have similar model parameters. Since model similarity among agents cannot be measured by a unified criterion in practice, the collaboration graph solved by fixed optimization is not always precise when local tasks vary. To fix this, we adopt algorithm unrolling and impose learnable similarity prior to make our graph learning model expressive and adaptive enough to handle various tasks. Experimentally, we validate our method on both regression and classification tasks and show that i) performance by collaboration learning is significantly better than solo-learning; ii) our unrolled graph learning method is consistently better than standard optimization.

Our contribution can be summarized as:
i) we formulate a mathematical framework for a human-like collaborative learning system, where each agent can autonomously build collaboration relationships to improve its own task-solving ability; and ii) we propose a distributed graph learning algorithm based on algorithm unrolling, enabling agents to find appropriate collaborators in a data-adaptive fashion.\\

\vspace{-0.3cm}
\section{Related works}
\label{sec:related works}
\noindent\textbf{Collaborative learning}.
Two recent collaborative learning frameworks achieve tremendous successes, including federated learning and swarm learning. Federated learning enables multiple agents/organizations to collaboratively train a model~\cite{2103.00710,9084352,rieke2020future,WarnatHerresthal2021SwarmLF,lalitha2018fully}. Swarm learning promotes extensive studies about multi-agent collaboration mechanism~\cite{jiang2017collaborative,1709.05412,minami2020knowledge,li2021learning,lei2022latency,hu2022where2comm}. In this work, we propose a human-like collaborative learning framework, emphasizing collaboration relationship inference, which receives little attention in previous works.

\noindent\textbf{Graph learning}.
Graph learning can be concluded as inferring graph-data topology by node feature\cite{8700659,8700665}. Typical graph learning (e.g. Laplacian inferring) can be solved by the optimization problem formulated by inter-nodal interaction\cite{1406.7842,1811.08790}. 
As traditional graph inference may fail when the objective cannot be well mathematically formulated, there are approaches applying graph deep learning models or algorithm unrolling~\cite{2206.08119,2110.09807,1912.10557,shlezinger2022model}. In this work, we leverage algorithm unrolling techniques to learn a graph structure, which combines both mathematical design and learning ability.

\section{Methodology}
\label{sec:method}

\subsection{Optimization problem}
\label{sec:problem}
Consider a collaboration system with $N$ agents. Each agent is able to collect local data and collaborate with other agents for an overall optimization. Let $X_i,Y_i$ be the observation and the supervision of the $i$th agent. The performance of a local model is evaluated by the loss $\LL_i(X_i,Y_i;\thet_i)$ ($\LL_i(\thet_i)$ for simplicity), where $\thet_i\in\mathbb{R}^M$ is the model parameter of the $i$th agent.
The pairwise collaboration between agents is formulated by a directed collaboration graph represented by the adjacency matrix $\mathbf W\in\mathbb{R}^{N\times N}$, whose $(i,j)$th element $w_{ij}$ reflects the collaboration weight from agent $i$ to $j$.
Note that we do not consider self-loops so the diagonal elements of $\mathbf W$ are all zeros.
Then the $i$th agent's partners are indicated by a set of outgoing neighbours with nonzero edge weights; that is, $\mathcal{N}_i = \{{\rm Agent}~j|w_{ij}>0\}$.
Inspired by social group-effect, agents are encouraged to find partners and imitate their parameters for less biased local model\cite{almaatouq2020adaptive}. Therefore, the global optimization is formulated as:
\begin{small}
\begin{align}
\label{eq:global_opt}
& \min_{ \{ \thet_i \}_{i=1}^N,\mathbf W} ~ 
\sum_{i=1}^{N}
\LL_i(\thet_i)+\lambda_1 \|\mathbf W\|_F^2
+\lambda_2\!\sum_{i,j=1}^N w_{ij}\|\thet_i\!-\!\thet_j\|^2_2\notag\\
&~~~~~~{\rm subject~to~}~\ \|\mathbf W\|_1=N, w_{ii}=0, w_{ij} \geq 0, \forall i,j,
\end{align}
\end{small}where $\lambda_1, \lambda_2$ are predefined hyperparameters. The first term reflects all the local task-specific losses;
the second term regularizes energy distribution of edge-weights; and the third term promotes graph smoothness;
that is, agents with similar tasks tend to have similar model parameters and have higher demands to collaborate with each other. 

However, in a distributed setting, there is no central server to handle the global optimization.  To optimize~\eqref{eq:global_opt} distributively, let $\mathbf w_i$ be the $i$th column of  $\mathbf{W}$ and we consider the following  local optimization for the $i$th agent:
\begin{align}
\label{eq:local_opt}
& \min_{ \thet_i,\mathbf w_i} \LL_i(\thet_i)
+\lambda_1 \|\mathbf w_i\|_2^2+\lambda_2 \sum_{j} w_{ij}\|\thet_i-\thet_j\|^2_2\notag
\\ 
&  {\rm subject~to~}~\|\mathbf w_i\|_1=1, w_{ii}=0, w_{ij} \geq 0, \forall j.
\end{align}
Note that the feasible solution space of~\eqref{eq:local_opt} is a subset of the feasible solution space of~\eqref{eq:global_opt} because of the first constraint. 
Each agent has no perception of all the collaboration relationships and can only decide its outgoing edges.
To solve problem~\eqref{eq:local_opt}, we consider an alternative solution, where each agent alternatively optimizes its $\thet_i$ and $\mathbf w_i$. The overall procedure is shown in \textbf{Algorithm~\ref{alg}}, which contains two alternative steps:

\begin{algorithm}[t]
\caption{Collaborative learning for Agent $i$}
\label{alg}
\begin{algorithmic}
\renewcommand{\algorithmicrequire}{$\mathbf{Input:}$}
\renewcommand{\algorithmicrepeat}{$\mathbf{Repeat:}$}
\renewcommand{\algorithmicuntil}{$\mathbf{Until:}$}
\renewcommand{\algorithmicensure}{$\mathbf{Output:}$}
\REQUIRE
\STATE\textbf{Initialization:} 
$\thet_i^{(0)}\leftarrow {\arg\min_\thet\ }\LL_i(\thet)$
\FOR{$t\leftarrow 0\ \mathbf{to} \ T_1-1:$}
    \IF{$t\mod T_2=0$}
    \STATE\textcolor{blue}{\# \textit{broadcast and update $\mathbf w_i$}}
    \STATE Broadcast $\thet_i^{(t)}$ to all agents
    \STATE Get $\mathbf\Theta^{(t)}=[\thet_1^{(t)},...,\thet_N^{(t)}]$ from other agents
    \STATE $\mathbf w_i\leftarrow$ graph\_learning$(\mathbf\Theta^{(t)})$ by \textbf{Algorithm 2} or \textbf{3} 
    \ELSE
    \STATE\textcolor{blue}{\# \textit{only communicate with partners}}
    \STATE
    Send $\thet_i^{(t)}$ to partners $\mathcal{N}_i = \{{\rm Agent}~j|w_{ij}>0\}$
    \STATE
    Get $\mathbf\Theta^{(t)}_{\mathcal{N}_i}=\{\thet_j^{(t)}|w_{ij}>0\}$ from partners
    \ENDIF
    \STATE\textcolor{blue}{\# \textit{update parameters}}
    \STATE 
    $\thet_i^{(t+1)}\leftarrow {\arg\min_\thet\ }\LL_i(\thet)+\lambda_2\sum_{j\in \mathcal{N}_i} w_{ij}\|\thet-\thet_j^{(t)}\|^2$
\ENDFOR
\ENSURE$\mathbf\thet_i,\mathbf w_i$
\end{algorithmic}
\end{algorithm}

\textit{1) Graph learning.}
This step allows each agent to optimize who to collaborate with. Through broadcasting, the $i$th agent obtains all the other agents' model parameters $\{\thet_j^{(t)}|j\neq i\}$; and then optimizes its local relationships with others by solving the subproblem:
\begin{align}
\label{eq:graphobj}
&
\min_{\mathbf w_i}~ \lambda_1 \|\mathbf w_i\|^2_2 + \lambda_2 \sum_{j} w_{ij}\|\thet_i^{(t)}-\thet_j^{(t)}\|^2_2  
\\ \nonumber
&
{\rm subject~to~}~\|\mathbf w_i\|_1 = 1, w_{ii}=0, w_{ij} \geq 0, \forall j.
\end{align}

Since~\eqref{eq:graphobj} is a convex problem,  we optimize $\mathbf w_i$ via the standard dual-ascent method.
Let $f(\mathbf{w}_i) = \lambda_1\|\mathbf w_i\|^2_2 + \lambda_2\sum_{j} w_{ij}\|\thet_i^{(t)}-\thet_j^{(t)}\|^2_2$. The dual-ascent updating step is,
\begin{small}
\begin{align*}
\mathbf{w}_i &\leftarrow \arg \min_{\mathbf{w}_i}~
f(\mathbf{w}_i)+z (\mathbf{1}^\top\mathbf{w}_i-1)+\mathcal{P}_{\mathbf{w}_i>0},
\\
z~~& \leftarrow z + p\cdot(\mathbf{1}^\top\mathbf{w}_i-1),
\end{align*}
\end{small}where $p$ is the stepsize and $\mathcal{P}_{\mathbf{w}_i>0}$ is a barrier function to ensure $w_{ij} \geq 0$. Until the convergence, we obtain the $i$th agent's collaboration relationship for the next iterations to update $\thet_i$.
The detailed process is shown in \textbf{Algorithm~\ref{DA}}. 

\begin{algorithm}[h]
\caption{Dual-ascend for graph learning}
\label{DA}
\begin{algorithmic}
\renewcommand{\algorithmicrequire}{$\mathbf{Input:}$}
\renewcommand{\algorithmicrepeat}{$\mathbf{Repeat:}$}
\renewcommand{\algorithmicuntil}{$\mathbf{Until:}$}
\renewcommand{\algorithmicensure}{$\mathbf{Output:}$}
\REQUIRE
$\mathbf \Theta=[\thet_1,...,\thet_N]\in \mathbb R^{M\times N},p=stepsize$
\STATE$\mathbf{Initialization:}$
$\mathbf w_i \leftarrow (\mathbf1^{N-1})/(N-1), z \leftarrow 0$
\STATE$\mathbf{Ensure:}$
$w_{ii} = 0$
\STATE$\mathbf{d}_i=[\|\thet_i-\thet_1\|_2^2,...,\|\thet_i-\thet_N\|_2^2]$
\REPEAT
\STATE$\mathbf w_i  \leftarrow{\rm ReLU}\left(-\frac{\lambda_2\mathbf{d}_i + z}{2 \lambda_1} \right)$
\STATE${\rm diff} \leftarrow \mathbf1^{\top} \mathbf w_i - 1$; $z \leftarrow z + p \cdot {\rm diff}$
\UNTIL{Convergence}
\ENSURE$\mathbf w_i$
\end{algorithmic}
\end{algorithm}

\textit{2) Parameter updating.}
This step allows each agent to update its local model by imitating its partners. Given the latest collaboration relationships $\mathbf w_i$ and the partners' parameters, each agent obtains its new parameter $\thet_i^{(t+1)}$ by optimizing:
\begin{eqnarray}
\label{eq:parameter_update}
\min_{\thet_i}\ \LL_i(\thet_i)
+\lambda_2\sum_{j} w_{ij}\|\thet_i-\thet_j^{(t)}\|^2_2,
\end{eqnarray}
where $\thet_j^{(t)}$ is the $j$th agent's local model parameter in the $t$th iteration. To provide a general analytical solution, we could consider a second-order Taylor expansion to approximate the task-specific loss $\LL_i$; that is,
\begin{equation}
\LL_i(\thet_i) \approx \frac{1}{2}(\thet_i-\alpha_i)^\top \mathbf H_i(\thet_i-\alpha_i) + \LL_i(\alpha_i) ,
\end{equation}
where
$\alpha_i={\arg\min_\thet\ }\LL_i(\thet)$, which is solved by gradient descent in the initialization step, and $\mathbf H_i$ is the Hessian matrix of $\LL_i$ at $\alpha_i$. By this approximation, the objective is in the form of a quadratic function determined by $\alpha_i$ and $\mathbf{H}_i$. Then, the optimization problem~\eqref{eq:parameter_update} becomes quadratic and we can obtain its analytical solution:
\begin{small}
\begin{align}
\thet^{(t+1)}_i\!=\!
\big( \mathbf H_i\!+\!2\lambda_2\|\mathbf{w}_i\|_1\mathbf I \big)^{-1}
\big(\mathbf H_i \alpha_i\! +\! 2\lambda_2 \sum\nolimits_{j} w_{ij}\thet^{(t)}_j \big).
\label{localupdate}
\end{align}
\end{small}\indent
To reduce communication cost, each agent executes parameter update by~\eqref{localupdate} without changing neighbour for $T_2$ rounds; and then, each agent will recalculate the collaboration relationship and update its partners. $T_1$ and $T_2$ are set empirically as long as $\thet_i$ converges.

\subsection{Unrolling of graph learning}
\label{ssec:Unrolling}
The optimization problem~\eqref{eq:global_opt} considers the quadratic term of graph Laplacian to promote graph smoothness, which is widely used in many graph-based applications. However, it has two major limitations. First, the $\ell_2$-distance might not be expressive enough to reflect the similarity between two models. Second, it is nontrivial to find appropriate hyperparameter $\lambda_1$ to attain effective collaboration graph.
To address these issues, we propose a learnable collaboration term to promote more flexibility and expressiveness in learning collaboration relationships. We then solve the resulting graph learning optimization through algorithm unrolling.

\vspace{0.1cm}
Let $\mathbf{D}_i \in\mathbb R^{M\times N}$ be the $i$th agent's parameter distance matrix whose $(m,j)$th element is $(\mathbf{D}_i)_{mj}= (\thet_{im}-\thet_{jm})^2$ with $\thet_{im}$ the $m$th element of the $i$th agent's model parameter $\thet_{i}$. The original graph smoothness criteria can be reformulated as
$
{\bf 1}^T_M \mathbf{D}_i  \mathbf{w}_i \ = \  \sum_{m=1}^M \sum_{j=1}^N  (\thet_{im}-\thet_{jm} )^2 w_{ij},
$
where ${\bf 1}_M \in \mathbb{R}^M$ is an all-one vector. To make this term more flexible, we introduce trainable attention to reflect diverse importance levels of model parameters and reformulate the graph learning optimization as
\begin{small}
\begin{align}
\centering
& \min_{\mathbf w_i}\ \frac12\|\mathbf Q\mathbf D_i\mathbf w_i\|^2_2 =
\frac12
 \sum_{m=1}^M \bigg( \sum_{j=1}^N q_{m} (\thet_{im}-\thet_{jm} )^2 w_{ij} \bigg)^2
 \nonumber \\ 
 \label{eq:unrolling}
&\quad{\rm subject~to~}\|\mathbf w_i\|_1 = 1, w_{ii}=0, w_{ij} \geq 0, \forall j,
\end{align}
\end{small}where $\mathbf Q=\mathbf{diag}(q_1,..q_M)$ with $q_k>0$ reflecting the importance of the $m$th parameter.
The new objective merges the original graph smoothness criteria and  energy constraint. It is also quadratic to make the optimization easier. According to the proximal-descend procedure, when the stepsize is not so large, the optimizing iteration can be formulated as:
\begin{equation*}
\mathbf w^{k+1}_i \leftarrow {\rm ReLU} \left({\rm Proj}_{\mathcal{D}}(\mathbf I-\mu \mathbf D_i^\top\mathbf Q^2\mathbf {D}_i)\mathbf w_i^k \right),
\end{equation*}
where $\mu$ is the stepsize and projection ${\rm Proj}_{\mathcal{D}}$ is specified as:
$$
{\rm Proj}_{\mathcal{D}}(\mathbf V)\equiv \mathbf V-\frac{(\mathbf1_{M}\mathbf V-1)\mathbf1_M}{M}.
$$

To reduce parameter complexity, we use one diagonal matrix $\mathbf P=\mu\mathbf Q^2>\gamma$ to integrate $\mu$ and $\mathbf Q^2$.
The $m$th element $p_m$ on the diagonal of $\mathbf P$ can be interpreted as the stepsize made by the smoothness of the $m$th parameter and
$\gamma$ is a lower limit in training to avoid $p_m$ from degrading to 0.
$\mathbf{P}$ can be supervised by loss $\LL_P$ formulated by the average performance of all the agents with regard to actual local task.
The unrolled forwarding of $K$ iterations is showed in \textbf{Algorithm~\ref{graphalg}}. 
For more adaptability, the output is not necessarily the actual solution to~\eqref{eq:unrolling} , which means few iterations is needed and the output will be largely decided by $\mathbf P$.

\begin{algorithm}[H]
\caption{Unrolled graph learning}
\label{graphalg}
\begin{algorithmic}
\renewcommand{\algorithmicrequire}{$\mathbf{Input:}$}
\renewcommand{\algorithmicrepeat}{$\mathbf{Repeat:}$}
\renewcommand{\algorithmicuntil}{$\mathbf{Until:}$}
\renewcommand{\algorithmicensure}{$\mathbf{Output:}$}
\REQUIRE
$\mathbf \Theta=[\thet_1,...,\thet_N]^\top\in \mathbb R^{M\times N}$
\STATE$\mathbf{Initialization:}$
$\mathbf w^0_i = (\mathbf1^{N-1})/(N-1)$
\STATE$\mathbf{Ensure:}$
$w_{ii} = 0$
\FOR{$k\leftarrow 0\ \mathbf{to}\ K-1:$}
\STATE$\mathbf w_i^{k+1}={\rm ReLU[Proj}_{\mathcal{D}}(\mathbf I- \mathbf D_i^\top\mathbf P\mathbf D_i) \mathbf W^k]$
\ENDFOR
\STATE $\mathbf w_i=\mathbf w^K_i/\|\mathbf w^K_i\|_1$ \textcolor{blue}{\textit{\# normalize}}
\ENSURE$\mathbf w_i$
\end{algorithmic}
\end{algorithm}

\section{Experiments}
\label{sec:experiments}
\begin{table}[H]
\centering
\setlength{\tabcolsep}{3pt}
\begin{tabular}{ccccc}
\hline
\multicolumn{1}{l}{Task/Method} & \begin{tabular}[c]{@{}c@{}}No\\ Colla.\end{tabular} & \begin{tabular}[c]{@{}c@{}}Original\\ GL\end{tabular} & \begin{tabular}[c]{@{}c@{}}Unrolled \\ GL\end{tabular} & \begin{tabular}[c]{@{}c@{}}Fixed\\ Colla.\end{tabular} \\ \hline
\multicolumn{2}{l}{{\ul \textbf{Regression:}}}                                        & \multicolumn{1}{l}{}                                  & \multicolumn{1}{l}{}                                   & \multicolumn{1}{l}{}                                   \\
$L_{reg}$                       & 16.1854                                             & 3.5332                                               & 2.9642                                                 & 2.2294                                                 \\
GMSE                         & -                                                   & 1.7683                                                & 0.8056                                                & 0                                                      \\
\multicolumn{2}{l}{{\ul \textbf{Classification:}}}                                    & \multicolumn{1}{l}{}                                  & \multicolumn{1}{l}{}                                   & \multicolumn{1}{l}{}                                   \\
ACC                             & 0.6214                                              & 0.7268                                                & 0.7429                                                 & 0.7481                                                 \\
GMSE                         & -                                                   & 0.3115                                                & 0.1188                                                 & 0                                                      \\ \hline
\end{tabular}

\caption{Comparison on regression and classification tasks.}
\label{tab}
\end{table}
\begin{figure}[H]
    \begin{minipage}[b]{.49\linewidth}
      \centering
      \centerline{\includegraphics[width=\textwidth]{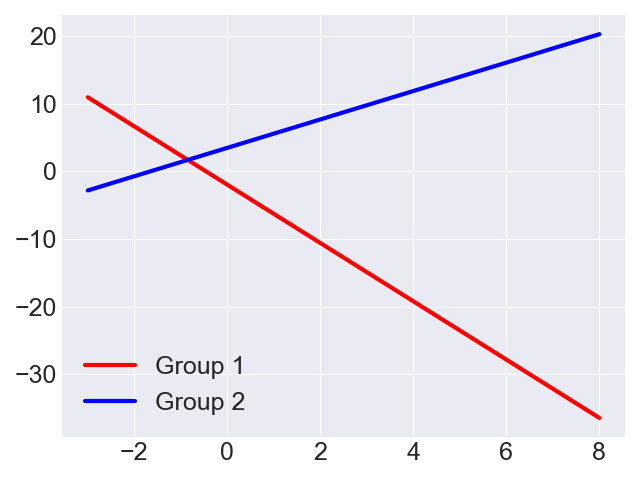}}
      \centerline{\small  (a) Ground-truth lines.}\medskip
    \end{minipage}
    \hfill
    \begin{minipage}[b]{0.49\linewidth}
      \centering
      \centerline{\includegraphics[width=\textwidth]{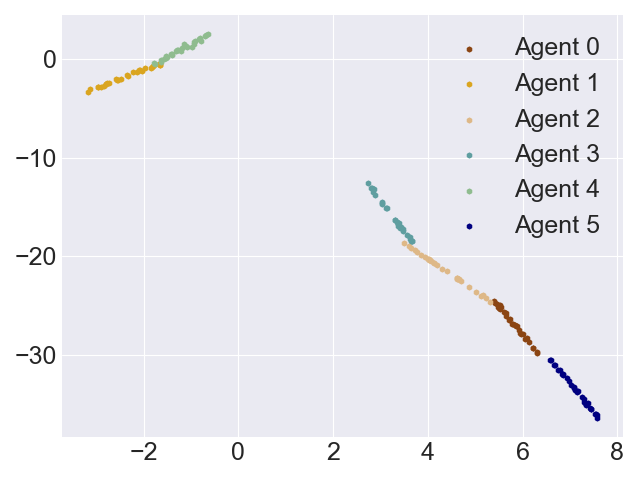}}
      \centerline{\small  (b) Noisy samples.}\medskip
    \end{minipage}
    \begin{minipage}[b]{0.49\linewidth}
      \centering
      \centerline{\includegraphics[width=\textwidth]{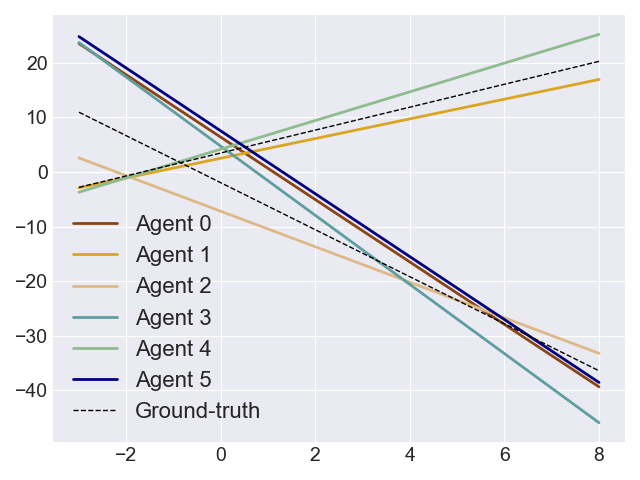}}
      \centerline{\small (c) Result without collaboration.}\medskip
    \end{minipage}
    \begin{minipage}[b]{0.49\linewidth}
      \centering
      \centerline{\includegraphics[width=\textwidth]{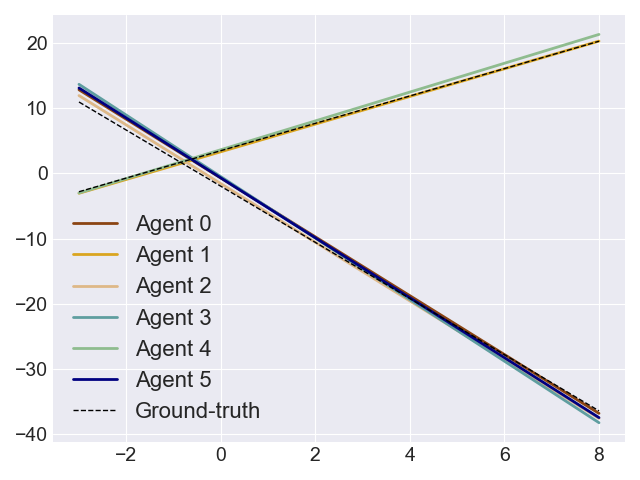}}
      \centerline{\small(d) Result by learned graph.}\medskip
    \end{minipage}
    \vspace{-0.3cm}
    \caption{A visualized example of data and result in regression.}
    \label{fig:regres}
\end{figure}

To validate our model, we design two type of local tasks (regression and classification) and compare the performance of our unrolled network with different collaboration schemes.

\subsection{Linear Regression}
\label{ssec:regression}
\textbf{Dataset.} We consider two different lines. Each agent only gets noisy samples from a segment of one line and aims to regress the corresponding line function, see \textbf{Fig.~\ref{fig:regres}}. 
To achieve better regression, each agent can collaborate with other agents and get more information about the line.
The challenges include: i) how to find partners that are collecting data from the same line; and ii) how to fuse information from other agents to obtain a better regression model. 

\vspace{0.1cm}
\noindent\textbf{Evaluation.}
We consider two evaluation metrics: one for regression and the other one for graph learning. Let the regression error be
$
L_{reg} = \frac1N\sum_{i=1}^N[(\widehat{k}_i-k_i)^2+(\widehat{b}_i-b_i)^2],
$
where $\widehat{k}_i, \widehat{b}_i$ are the line parameters estimated by the $i$th agent and $k_i, b_i$ are the ground-truth line parameters of the $i$th agent. 
The graph structure is evaluated by
$
{\rm GMSE} = \frac1N\sum_{i=1}^N\|\mathbf{\widehat w}_i-\mathbf{w}_i\|_F^2,
$
$\mathbf{\widehat w_i}$ is the estimated edge-weights and $\mathbf{w}_i$ is the ground-truth edge-weights, where the edge-weights are uniformly distributed only among agents in the same task group fitting the same line.

\vspace{0.1cm}
\noindent\textbf{Experimental setup.}
We compare four methods: i) local learning without collaboration; ii) collaboration by the predefined ground-truth graph; iii) collaboration by original optimization \textbf{Algorithm~\ref{DA}} with well-tuned $\lambda_1$; and iv) collaboration by unrolled model \textbf{Algorithm~\ref{graphalg}}. We set the same $\lambda_2$ for all methods to ensure fairness. The unrolled model is pretrained on a training set and the hyperparameter $\mathbf{P}$ is supervised by the regression error $L_{reg}$. Then all models are tested on the same testing set including different partitions and data.

\vspace{0.1cm}
\noindent\textbf{Results.} 
\textbf{Table~\ref{tab}} shows that i) collaboration brings significant benefits; ii) unrolling works better than pure optimization; iii) the unrolled graph is closer to the ground-truth graph. 
These results are expected because $\thet_i=(k_i,b_i)^\top$ is non-Euclidean and the unrolled model can learn a more suitable evaluation than $\ell_2$-distance.
\textbf{Fig.~\ref{fig:regres}} visualizes the regression results, which reflects the consistent patterns with Table~\ref{tab}.

\begin{figure}[t]
\begin{minipage}[b]{0.51\linewidth}
  \centering
  \centerline{\includegraphics[width=\textwidth]{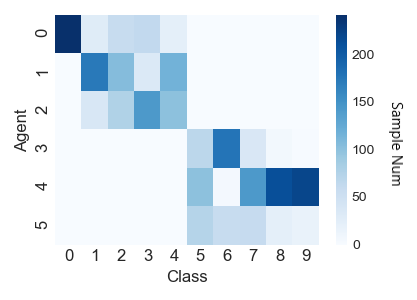}}
\end{minipage}
\hfill
\begin{minipage}[b]{0.48\linewidth}
  \centering
  \centerline{\includegraphics[width=\textwidth]{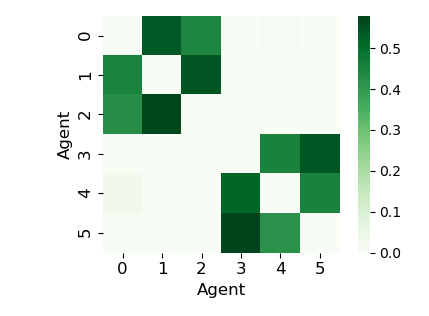}}
\end{minipage}
\vspace{-0.7cm}
\caption{Data distribution (left) and the learned collaboration weight matrix $\mathbf{W}$ (right) in one classification task.}
\label{fig:classdata}
\end{figure}

\vspace{-0.1cm}
\subsection{Classification}
\label{ssec:classification}
\textbf{Dataset.}
The dataset we adopt is a reduced MNIST. Original $28\times28$ grid is processed by a pre-trained ResNet and reduced by PCA to a vector of $\mathbb{R}^{20}$. There are 10 types of samples. 
Agents are divided into two groups: group 1 classify type 1-5 and group 2 classify type 6-10.
The samples are non-IID, see \textbf{Fig.~\ref{fig:classdata}}.
The challenges also include how to find partners having the same data category and how to fuse information for less biased perception.

\vspace{0.1cm}
\noindent\textbf{Evaluation.}
Similarly, there are two evaluation metrics:
the classification performance is evaluated by agents' average ACC  and graph learning is evaluated by $\rm GMSE$.

\vspace{0.1cm}
\noindent\textbf{Experimental setup.}
The baselines and testing process are the same as \ref{ssec:regression}.
Differently, the local model at each agent is a linear classifier for 5 classes and $\LL_i(\thet_i;X_i,Y_i)$ is defined as the cross-entropy loss function. 
Because there is no ground-truth local parameter,
\vspace{0.05cm}
in pretraining the unrolled hyperparameter $\mathbf{P}$ is supervised by $\LL_P=\sum_{i=1}^N\LL_i(\thet_i;X_i,Y_i)$.

\vspace{0.1cm}
\noindent\textbf{Results.}
The results are also shown in \textbf{Table~\ref{tab}}.
Note that the unrolled model gains a more obvious enhancement than \ref{ssec:regression} as local parameter has higher dimension.
A visualized example of the learned matrix $\mathbf{W}$ is shown in \textbf{Fig.~\ref{fig:classdata}}. 
\vspace{-0.1cm}
\section{Conclusion and future work}
\label{conclusion}
\vspace{-0.1cm}
We proposed a distributed multi-agent learning model inspired by human collaboration and an unrolled model for collaboration graph learning. By experiments in different tasks, we verify that: i) our human-like collaboration scheme is feasible; ii) our unrolled graph learning can improve performance in various tasks.
Currently, the local tasks in our experiments are rudimentary trials.  
In future works, we will apply our framework to more complicated nonlinear local models for more versatility.
\bibliographystyle{IEEEbib}
\bibliography{ref}

\newpage
\section*{Appendix}

\begin{table*}[t]
\centering
\setlength{\tabcolsep}{5mm}
\renewcommand\arraystretch {1.2}
\begin{tabular}{c|cc|cc}
\hline
Task                                               & \multicolumn{2}{c|}{Regression} & \multicolumn{2}{c}{Classification} \\ \cline{2-5} 
\begin{tabular}[c]{@{}c@{}}/\\ Method\end{tabular} & $L_{reg}~(\downarrow)$      & GMSE $(\downarrow)$        & ACC $(\uparrow)$            & GMSE $(\downarrow)$         \\ \hline
No Colla. (Lower bound)                                          & 16.1854          & -            & 0.6214           & -               \\
\hline
Graph Lasso\cite{Mazumder2011TheGL}                & 7.6042           & 4.6187       & 0.6787           & 1.1829          \\
L2G-ADMM\cite{2110.0980}                          & 3.9842           & 1.6916       & 0.7166           & 2.0248          \\
Unrolled GL (Ours)                                        & 2.9642           &  0.8056       & 0.7429           & 0.1188          
\\ \hline 
Fixed Colla. (Upper bound)                                        & 2.2294           & 0            & 0.7481           & 0               \\ \hline
\end{tabular}
    \caption{Comparisons with two additional baselines}
    \label{tab}
\end{table*}

\subsection*{A. Comparison with other baselines}
To make a more comprehensive comparison, we also introduce two additional baselines: Graphical Lasso and L2G-ADMM.
Graphical Lasso is a classical graph learning algorithm for undirected Gaussian graphical model~\citesec{Mazumder2011TheGL}.
L2G-ADMM is a model-based graph Laplacian learning method solved by ADMM~\citesec{2110.0980}.
Note that both of the two baselines are centralized methods.

The results are shown in \textbf{Table~\ref{tab}}. We can see that the proposed unrolled method significantly outperforms two baselines. That is because our distributed model has regularization for each column for $\mathbf W$, which can encourage each agent to refer to others' parameters.
By contrast, L2G-ADMM does not have regularization specifically designed for collaboration.
Graphical Lasso is not reliable when the number of local parameters $M$ is small ( local model parameters are $M$ samples of $N$-dimensional Gaussian distribution).

\subsection*{B. Parameter settings in our experiments}
The settings of critical parameters in the experiment are listed below:
\begin{itemize}
    \item To ensure the same collaboration weight, we set $\lambda_2=0.1$ for all the collaborative methods.
    \item In regression tasks $\lambda_1=3$ and in classification tasks $\lambda_1=0.05$ (tuned by grid search on the training set).
    \item In both tasks, the unrolling steps $K=10$.
    \item To compare the performance under limited collaboration times, in both tasks $T_1/T_2=2$, which means the agents can change collaborators for 2 times.
    We set $T_2$ large enough to ensure the convergence (10 for regression and 200 for classification). 
\end{itemize}

In regression tasks, each agent gets 100 sample points.
In classification tasks, each agent gets about 250 samples.
The local samples for each agent will not change in one test.

\subsection*{References}
\begingroup
\renewcommand{\section}[1]{}
\bibliographystylesec{IEEEbib}
\bibliographysec{ref2}
\endgroup
\end{document}